\documentclass[journal]{IEEEtran}

\ifCLASSINFOpdf

\usepackage{hyperref}
\usepackage{url}
\usepackage{caption}
\usepackage{marvosym}
\usepackage{amsmath}

\usepackage{enumitem}
\usepackage{subfigure}
\usepackage[english]{varioref}
\usepackage[noabbrev,english]{cleveref}
\usepackage{cite}
\usepackage{url}
\usepackage{tikz}
\usetikzlibrary{shapes,arrows}
\usetikzlibrary{calc,patterns,decorations.pathmorphing,decorations.markings}
\usetikzlibrary{intersections}

\usepackage[subnum]{cases}
\usepackage[export]{adjustbox}
\usepackage{tikz}
\tikzset{>=latex}
\usepackage{rotating}

 \usepackage{cite}
 \usepackage{url}
 \usepackage{tikz}
 \usetikzlibrary{shapes,arrows}
 \usepackage[european]{circuitikz} 
 \usepackage{enumitem}
 
\usepackage{bm}
\usepackage[english]{varioref}
\usepackage[noabbrev,english]{cleveref}
\usepackage{tikz}
\usetikzlibrary{patterns,calc}
\usepackage{comment} 
 \usepackage{rotating}
 \usepackage{tikz}
 \usepackage{tikz}
 \usetikzlibrary{shadows}
 \usetikzlibrary{matrix, positioning}
 \usepackage[squaren]{SIunits}
 \usepackage{graphicx} 
  \usepackage{graphicx}
  \usepackage{cleveref}
  \usepackage[subnum]{cases}
  \usepackage[export]{adjustbox}
  \usepackage{tikz}
  \usepackage{amssymb}
  \usepackage{pifont}
  \usepackage{marvosym}
\usepackage{balance}
\usepackage{multicol}
\usepackage{flushend}

\usepackage{graphicx,subfigure}
\usepackage{makecell}
\usepackage{setspace}

\usepackage[T1]{fontenc}

\usetikzlibrary{fit}
\usepackage{adjustbox}

\usepackage{array}
\begin{document}

\newcommand{\svdots}{\raisebox{3pt}{$\scalebox{.15}{\ddots}$}} 
\newcommand{\sddots}{\raisebox{3pt}{$\scalebox{.35}{$\ddots$}$}}

\newcommand{\mC}[2]{\multicolumn{1}{ N{#1}|}{#2}}
\newcommand{\mD}[2]{\multicolumn{1}{|N{#1}|}{#2}}

\newcolumntype{P}[1]{>{\centering\arraybackslash}p{#1}}
\newcolumntype{M}[1]{>{\centering\arraybackslash}m{#1}}



\title
{Metarobotics for Industry and Society: \\
Vision, Technologies, and Opportunities}

%

\author{Eric Guiffo Kaigom
\thanks{Eric Guiffo Kaigom is with the  Faculty of Computer Science and Engineering, Frankfurt University of Applied Sciences, Nibelungenplatz 1,
	60318 Frankfurt/Main, Germany. Contact: kaigom@fb2.fra-uas.de
	
	© 2024 IEEE. Personal use of this material is permitted. Permission from IEEE must be obtained for all other uses, in any
current or future media, including reprinting/republishing this material for advertising or promotional purposes, creating new
collective works, for resale or redistribution to servers or lists, or reuse of any copyrighted component of this work in other
works.
\vspace{0.1cm}

Published on IEEE Transactions on Industrial Informatics, Volume 20, Issue 4, April 2024, DOI: \url{https://doi.org/10.1109/TII.2023.3337380}
}%
} 
\maketitle

\begin{abstract} 
Metarobotics aims to combine next generation wireless communication,  multi-sense  immersion, and collective intelligence   to  provide  a pervasive, itinerant, and non-invasive access and interaction with distant  robotized applications. Industry and society are expected to benefit from these functionalities. For instance, robot programmers will no longer travel worldwide to plan and test robot motions, even collaboratively. Instead, they  will have a personalized access to robots and their environments from anywhere, thus spending more time with  family and friends. Students enrolled in robotics courses will be taught under authentic industrial conditions in real-time.  This paper describes objectives of Metarobotics in  society, industry, and in-between. It  identifies and   surveys technologies likely to enable their completion and provides an architecture  to put forward the interplay of key components of Metarobotics. Potentials for self-determination, self-efficacy, and work-life-flexibility in robotics-related applications in \mbox{Society 5.0, Industry 4.0, and Industry 5.0 are outlined.}

\end{abstract}

\begin{IEEEkeywords}
Robotics, Digital Twins, Metaverse, Collective Intelligence, 6G,  Holoportation, Industry 4.0$\slash$5.0, Society 5.0
\end{IEEEkeywords}

%
\IEEEpeerreviewmaketitle

	\section{Introduction}\label{intro1}
	Enhancing operational efficiency in  personalized production through smart decentralized robotized automation that quickly adapts to varying market conditions is  a key objective of the industry. 
	Efforts toward this end have culminated in the \textit{Industry 4.0} vision~\cite{prashar2023role,mourtzis2021design}. 
	 By contrast, improving the well-being and experience of workers on top of   
	Machine-to-Machine (M2M) communication is at the heart of   \textit{Industry 5.0}~\cite{xian2023advanced}. It instills a human-centered,  intellectual, social, and ethical acumen into the industrial worklife that reaches out to a digital transformation- and service-driven comfort, resilience, and self-fulfillment  of citizens  \mbox{as core pillars of  \textit{Society 5.0}~\cite{carayannis2023dark}.}

However, there is a widening  gap between 
physical  human-robot-collaboration (HRC) restricted to  factories or home settings and  personal as well as professional  expectations of citizens. The generation  Z of workers born after 1997 is an example. New work practices preferred by this generation are driven by  digitalization, smart and boundless mobility, along with interconnected  decisions. Forbes mentions that \textit{"...generation Z are spending more time in Metaverse-related scenarios and have a closer relationship with their online selves than any generation prior."}~\cite{Forbes}. Since such features and functionalities are barely integrated in HRC, societal ramifications and  {implications of the gap become apparent.}

    \begin{figure}[t]
	\includegraphics[width=0.85\columnwidth]{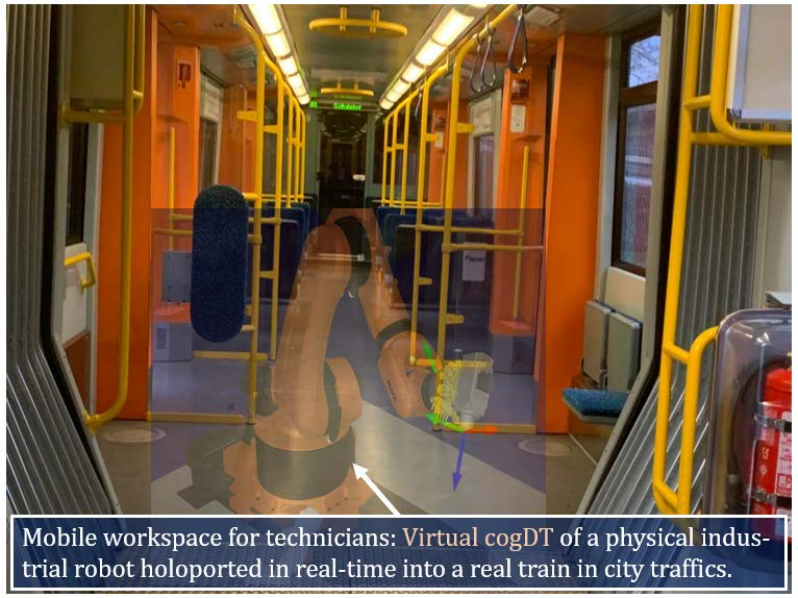}
	\centering
	\caption{Pervasive and itinerant HRC in a mobile workspace.}
	\label{basicidea}
\end{figure}

~An approach to fill in this gap is to take advantage of the increasing adoption  of immersive interfaces, including virtual$\slash$augmented reality (VR$\slash$AR) glasses, by citizens. This can be supplemented  with wireless data transfer beyond Ultra Reliable Low Latency Communication (b-URLLC), such as 6G, and collective intelligence based upon  jointly reasoning \text{cognitive Digital Twins} (cogDTs) to constantly ensured HRC \cite{mourtzis2022closed}. The seamless  appearance, disappearance, and multi-modal interaction with  cogDTs of   physical robots  and their environments spatially teleported (i.e., \text{holoported}) into  physical locations, even in motion,  enables a contextualized access and operation of robots  available on demand anytime and anywhere (see \cref{basicidea}). 
Citizens can  leave usual living- and workspaces while collaboratively  carrying out  robotized applications from \textit{wherever}, \mbox{as shown in the smart  mobility layer of	\cref{threelayer}.}

This novel form of \underline{p}ervasive and \underline{i}tinerant HRC termed as pi-HRC or \text{$\pi$-HRC} enables flexible, mobile, and sustainable workspaces, while  strengthening inclusion and resilience. Besides the reduction of the carbon footprint, and thus the ecological impact of workers free to work from any location, social distancing is self-regulated while maintaining  interactions with distant robots and collaborators. In-house and outsourced employees  work  with customers on distant robotized applications as if they were all physically, technically, and socially co-present in the same remote  place, even during pandemics.

Since collaborators have their own six dimensional (6D) views  of shared   cogDTs with depth, color, and light perception,  limitations of a fixed 2D  video-streaming are  overcome in robotized applications. For instance,  workers can remotely walk around the same holoported workpiece to teach target robot poses and thereby jointly specify goals. This accelerates  co-programming and facilitates socialization among workers despite distance. 
 Also, the potential for decent and upskilling work conditions, especially under hazards, such as  heat or smoke exposition in glass factories~\cite{barker2022collaborative}, 
are incentives for  citizen scientists and professional researchers to contribute to a shared and evolving knowledge base  that informs decisions empowering citizens in $\pi$-HRC. 
The  \textit{"human touch"} for robotized \textit{"mass personalization"} advocated in \cite{ostergaard2018welcome}  becomes actionable \textit{beyond} industrial borders to reach out to human-centered society at large.%
~This \textbf{\textit{Metarobotics}} can be realized  in a way that is location-agnostic, globally knowledge-driven, trustworthy, \mbox{uplifting for  workers, and noble for citizens.}
 
 \section{Contributions}\label{contribution}
\begin{figure}[t]
	\includegraphics[width=0.917\columnwidth]{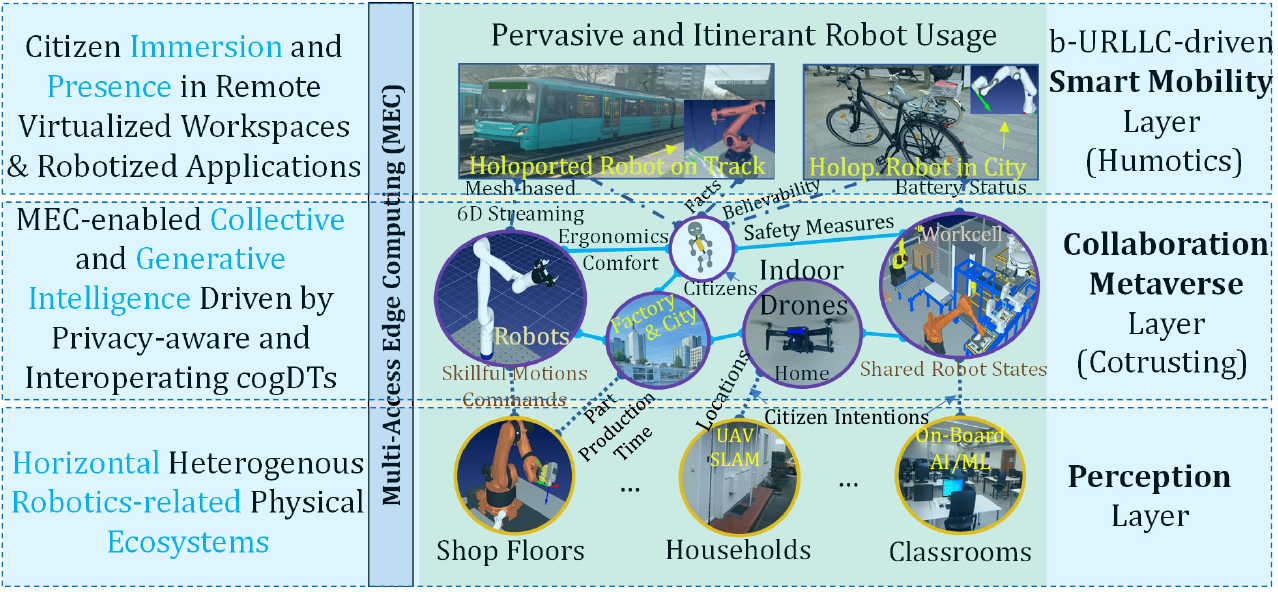}
	\centering
	\caption{A three layer view  on \textit{Metarobotics}.
	}
	\label{threelayer}
\end{figure}

\textit{Metarobotics} is introduced as a concept for the pervasive and itinerant interaction with distant robotized applications and their   enrichment with collective intelligence. Emerging technologies toward this end are identified and a survey thereof is provided. Discussed functionalities align with  goals pursued by Industry $4.0$, Industry $5.0$, and Society $5.0$ by fostering 
 \begin{itemize}
 	\item  a \underline{hu}man-centered \underline{mo}bility of robo\underline{tics} beyond industry and society, called \textit{Humotics}, in the top layer of \cref{threelayer}. It  transcends  distance,  perpetuates  a flexible proximity between robots and citizens, and facilitates  inclusion. Mobile workspaces and operational efficiency  {are likely to   benefit from  this agility to anticipate and adapt course of actions and enhance comfort in robotized applications.} 
 	\item a \underline{co}llective  \underline{trust}ed \underline{in}telli\underline{g}ence  termed as \textit{Cotrusting} in the middle layer of \cref{threelayer}. It  goes beyond individual capabilities of  citizens and robots. Their networked and interoperating cogDTs fed with industrial and societal data (see bottom layer in \cref{threelayer}) seamlessly augment citizens with otherwise hidden  information and knowledge that they transform into competitive advantages to sustainably  \mbox{achieve  robotized tasks while preserving privacy.} 
 \end{itemize}

 \begin{figure*}[ht!]
	\centering
	\includegraphics[width=1.6\columnwidth]{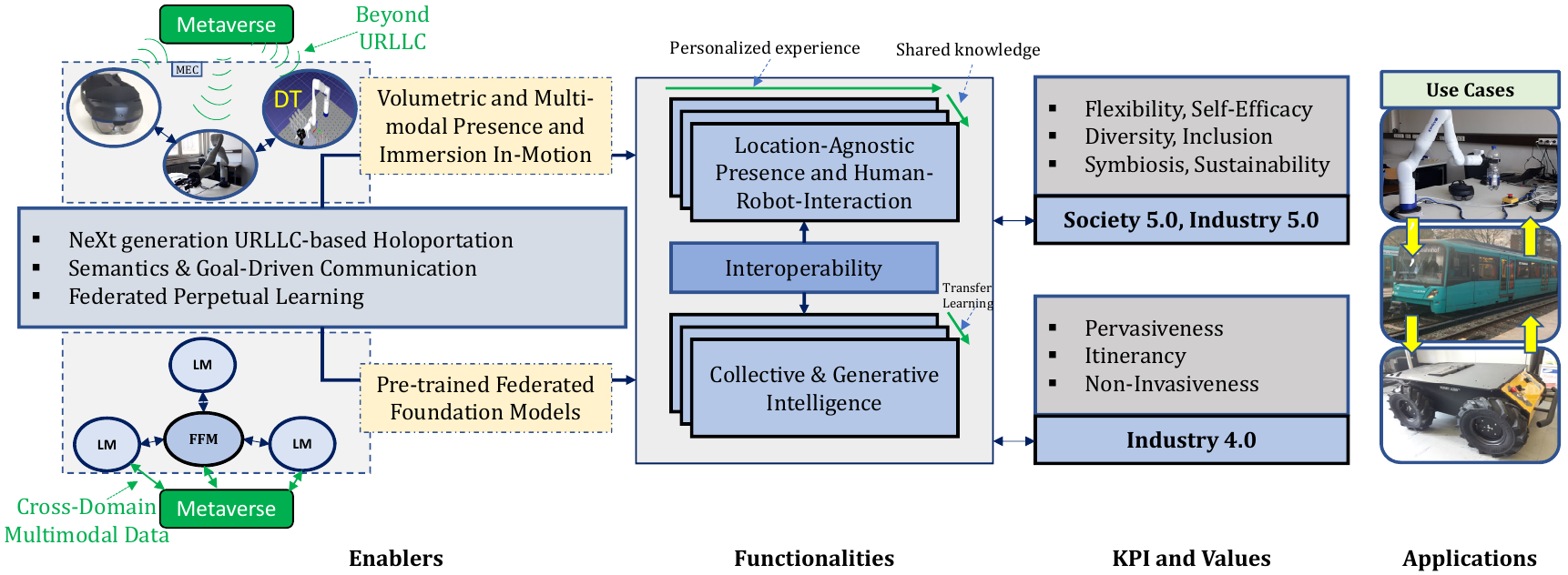}
	\caption{Enablers, Functionalities, Key Performance Indexes (KPI), and envisioned examples of use cases of \textit{Metarobotics}.}
	\label{metarobotics}
\end{figure*}

The paper is structured as follows. \Cref{relatedwork} describes  how \text{Metarobotics} extends related work. Core objectives of \text{Metarobotics} are introduced in \cref{Metarobotics}. 
\Cref{enablingtechnologies} surveys technologies  leveraged by \text{Metarobotics} and combined in its architecture to meet these objectives. Use cases populate the paper to exemplify the  practicability of these technologies. Finally, \mbox{\cref{conclusioon} concludes the paper with further challenges.}
 
 \begin{table}[t!]
 	\centering
 	\begin{center}
 		\resizebox{\columnwidth}{!}{%
 			\begin{tabular}{ |c||c|c ||c | }
 				\hline
 				\textbf{KPI}&\textbf{5G}&\textbf{6G}& \textbf{Enabling Technologies}\\
 				\hline 
 				\hline
 				Capacity 	&\makecell{10MBps\\ $/m^2$}&\makecell{$10^3\times$\\ $/m^3$}&\makecell{Visible light and Terahertz spectrum,\\Spatial multiplexing, etc.}\\
 				\hline
 				Downlink 	&20GBps&50$\times$ &\makecell{Non-Orthogonal Multiple Access (NOMA) \\
 					multiple-input multiple-output (MIMO), etc.}\\
 				\hline
 				Uplink 	&10GBps&$100\times$&\makecell{Non-Orthogonal Multiple Access (NOMA) \\
 					multiple-input multiple-output (MIMO), etc.}\\
 				\hline
 				Latency	&1 ms&$0.1\times$&\makecell{Multi-access Edge Computing (MEC), Caching, \\ Semantics-based individualization of communication,\\
 					Reinforcement Learning (RL) $\slash$  Deep Learning (DL)-based \\   allocation of resources, intelligent coding/decoding, etc. }\\
 				\hline
 				Reliability&$10^{-5}$&$ 10^{-4}\times$&\makecell{Multi-hop, Intelligent
 					Reconfigurable Surfaces (IRSs),\\Edge AI, RL-based robustification }\\
 				\hline
 				\makecell{User\\ experience}	&\makecell{50MBps\\ (2D)}&\makecell{$200\times$\\ (3D)}&\makecell{MEC, Terahertz spectrum, NOMA, context awareness,\\ Semantics-based personalization of communication,\\ IRSs,   Ambient Backscatter Communication (ABC), etc. }\\
 				\hline
 				Mobility	&500km/ h&$2\times$&Space-Air-Underwater-Terrestrial Integrated Networks\\
 				\hline
 				\makecell{Sensing\\Localization}	&\makecell{10 cm\\ (2D)}&\makecell{$ 0.1\times$\\ (3D)}& \makecell{Terahertz and mmWave spectrum,\\ SiGe BiCMOS, Bulk and PD-SOI CMOS, etc.}\\
 				\hline
 				\makecell{Energy/bit\\Efficiency}	&-&1 pJ/bit& \makecell{Edge Computing, IRS, ABC, Energy Harvesting, \\ Simultaneous Wireless Information \& Power Transfer,\\ Wireless Power Transfer (WPT), Network Functions\\  Virtualization (NFV), Software-Defined Networking (SDN),\\ Deep Reinforcement Learning-based efficient sleep/wakeup \\dynamics, Optical Wireless Communication (OWC), etc.
 				} \\ 
 				\hline
 			\end{tabular}
 		}
 	\end{center}
 	
 	\caption{5G $\slash$ 6G comparison. Numerical KPIs from \cite{strinati20216g}.}  
 	\label{table:v4v5}
 \end{table}

    \section{Related Work}\label{relatedwork}
 Enabling technologies behind  \textit{Humotics} and \textit{Cotrusting}, including  b-URLLC, Holoportation,  cogDTs, and Multi-Access Edge Computing (MEC), have been surveyed so far either  independently from each other or in a generic way~\cite{tataria20216g}. By contrast, this work surveys these technologies while emphasizing on their combination shown in \cref{overall,metarobotics,threelayer,robotaas} to meet the goals of  \textit{Metarobotics} introduced in section \ref{contribution}.~%
In \textit{Humotics}, for instance, b-URLLC is expected to achieve an unnoticeable transmission  of  holoported assets in terms of latency, throughput, and reliability from the edge to any remote location to propel actionable cogDTs, as pointed out in the MEC-layer of \cref{overall,robotaas}.  Semantics, goal orientation, and online learning behind 6G are thus leveraged by   \text{\textit{Metarobotics}}-based $\pi$-HRC for less transmitted data, reduced energy consumption, and more bandwidth~\cite{strinati20216g}. This contrasts with~\cite{SHI20221867} that exploits 5G  for HRC. In fact,   limitations of 5G (see  \cref{table:v4v5}) might impede $\pi$-HRC. Similarly, \textit{Cotrusting} harnesses MEC-enabled blockchained transactions to preserve privacy.%
 
A few surveys focusing on Holoportation have pointed out the importance of multi-modal feedback to raise the Quality of Experience (QoE) of citizens. The sense of touch is mentioned in e.g. \cite{petkova2022challenges} and  ISO/IEC 18039:2019. However, advantages of 6G networks for  ultra-massive {personal-level}  support are not considered. Further modalities, such as a thermal sensation and visual deformation feedback, were omitted in surveys. Yet, these sensations are crucial for  robotized tasks and predictive maintenance (e.g., situation awareness related to heating in the glass industry), as well as social interactions (e.g., touching a  coffee cup, handshaking), \mbox{and thus included in this paper.} 
 
 A challenge around multi-modal feedback in Holoportation is the synchronization of independent information channels. In this paper, a survey of  recent advances in synchronization techniques for multi-modal transmissions is provided. Achieving multi-modal feedback also offers  opportunities to technically and socially engage, for example, the generation Z in  professional and personal robotized applications, following thereby core objectives of Industry 4.0 and Industry 5.0.%

 A  state-of-the-art review of human-centered HRC has been recently proposed  in \cite{semeraro2023human}.%
 ~Beyond human interpretations in HRC and striving for human satisfaction in Society 5.0, the survey in this paper emphasizes  on  robot centricity. This contributes to symbiosis  and \mbox{shared autonomy \cite{selvaggio2021autonomy} in  $\pi$-HRC.} 
 
This paper conceptualizes the usage of b-URLLC-based and multi-sense Holoportation for  $\pi$-HRC that benefits  citizens in industry, society, and in-between. It summarizes  how \text{\textit{Metarobotics}}  capitalizes on surveyed  technologies in an architecture (see \cref{overall}) for that.  To the best of our knowledge, we are not aware of any previous work that provides such a recent \mbox{integration of  emerging technologies toward this end.} 
 
  \section{Metarobotics}\label{Metarobotics}
 
\subsection{Motivation}
\textit{Metarobotics} arises from the need to develop a technology-mediated and human-centered framework  that fosters self-determination, efficacy, and comfort in robotics-related applications. Self-determination conceptualizes  motivation around three  pillars which  are competence, autonomy, and inclusion \cite{gagne2022understanding}. Hence, citizens can perceive and understand challenges through multiple modalities, define personal and professional goals, and freely design  solutions under a transparent but holistic assistance of interconnected cogDTs. This assistance is provided  in a uniform transition between society and industry that considers current contexts and elevates  capabilities of citizens as well as  incorporates  global constraints  and shared intelligence. Finally, efforts of empowered citizens  materialize themselves in the \mbox{completion of robotized tasks in remote areas.} 

These objectives can be achieved in the  top layer of \cref{threelayer} regardless of  smart mobility modes of citizens. They are given access to virtualized  applications and empowered  with facts inferred in the middle layer using structured information on remote physical siblings of applications run  in the bottom layer. Conversely,  IoT devices and citizens return information  on the mobile workspace (e.g., ambient lighting, battery status) and appreciations of the trustworthiness of inferred facts to  enhance the believability of  knowledge from the middle  layer of the collaboration Metaverse. Here, the notion of collaboration  is disentangled from  usual restrictions to real humans. Collaboration agents encompass embodied avatars and  cogDTs. They semantically interoperate and reason in the Metaverse with a  depth and breadth of information gained across domains in the  bottom layer in \cref{threelayer}, along with  privacy-preserving learning and  attention skills that outperform the capabilities of  single humans they augment in terms of e.g. uncertainty handling and informed decision support. CogDTs are loosely coupled with their physical surrogates (e.g., a robot) that  they mirror, monitor, and \mbox{control to achieve goals in the bottom layer.} 

\subsection{Definition}
\textit{Metarobotics} is a software-defined framework (SDF) that strives to enable a  location-independent and continuous proximity as well as assisted interaction with robotized applications beyond traditional scopes and boundaries of society and industry (hence, the prefix "Meta-"). The softwarization enables a  reconfigurable availability  of   applications across  heterogeneous platforms, domains, and devices. 
 Standardized interfaces and information models, together with a granular control of atomic  (micro)services adapted on-demand~\cite{rath2023microservice}, facilitate the development of scalable, differentiated, and multidisciplinary solutions. Remote robotized applications can be   holoported into  collaboration spaces of the Metaverse using e.g. mesh- or point cloud-based reconstructions  and virtually  projected onto existing or prospective environments  without affecting the milieu.   Hence, \textit{Metarobotics} will  be non-invasive. In fact, cross-domain projections are digital and  do not modify augmented physical environments (see \cref{basicidea}). Each  projection ubiquitously enriches  the smart  mobility of citizens with spatial contexts and global cogDT-supported information about robotized applications. Additionally, the projection functionality engages citizens because virtualized applications (including  human partners, assets, and collaboration processes) can be \textit{spatially} (i.e., in 6D) up- or downscaled to comfortably and purposefully fit  in various environments like a confined cabin of an electric train in motion. Finally, each projection is sustainable by design through b-URLLC driven development, as  \mbox{mentioned in \cref{metarobotics,threelayer} and emphasized in \cref{table:v4v5}.} 
 
 Using b-URLLC to  combine these three dimensions (i.e., human-centric collective creation, sharing, and consumption of robotics-related knowledge, smart mobility,  remote and spatial access to robotized applications) in the layers of \cref{threelayer} distinguishes \textit{Humotics} from  approaches that only support single means of transport and 2D interfaces (e.g., tablets). Since \textit{Metarobotics} is comparatively characterized by an ultra-dense and heterogeneous mobility of citizens  with an on-demand and reliable proximity to remote robotized applications, as pursued by \textit{Humotics},  it needs to accommodate requirements set to network capacity and energy budget (see \cref{table:v4v5}). One approach to  achieve this goal is to leverage non-orthogonal multiple access (NOMA), millimeter Wave (mmWave), and Terahertz (THz) channels propelled by a base station caching and an application-dependent selection of channels \cite{jain2023dynamic}.  Channels for signal transmission of the downlink  mentioned in \cref{table:v4v5}  are assumed to be  adaptively selected as a function of e.g. the transmission rate requirement of the application (e.g.,  1 Tbps for Holoportation is associated with a Terahertz  channel) in \textit{Metarobotics} to enhance the spectral efficiency, still following \cite{jain2023dynamic}. Ultra-low latency, high data rates, and reliability of 6G  mentioned in \cref{table:v4v5} align with the Quality of Service (QoS)  required by \textit{Metarobotics} to control the motion  and force-sensitiveness  of most robots  in  closed feedback loops. Previously mentioned QoS properties also contribute to a high-fidelity multi-modal sensory (e.g.,  tactile) feedback for e.g. kinesthetic  guidance tasks in robotized applications. Furthermore,  the  QoE of citizens in terms of presence and immersion  will benefit from the QoS in \textit{Metarobotics}. Presence encompasses engaging multi-modal interactions with  virtualized environments in which  applications take place while immersion reflects a sensation of being completely located inside  \mbox{holoported environments \cite{mahmoud2023survey}.}  Localization in the Terahertz channel with multipath resolvability \cite{akyildiz2022terahertz} is an essential feature of  \textit{Metarobotics} for e.g. logistics use cases (see r.h.s of \cref{metarobotics}). Intelligent Reconfigurable Surfaces (IRSs) with known poses act as passive and adjustable base stations with low energy consumption \cite{chen2022tutorial} that are used for solid and compatible multipaths to localize assets like mobile robots in even harsh \mbox{environments (e.g., blockages of line-of-sight)  \cite{wymeersch2020radio}.}

As a SDF, \textit{Metarobotics} further targets a QoE that reflects how a configurable level of autonomy and efficacy  individualy culminates in satisfaction and comfort. For instance, the information overload on graphical user interfaces can be dynamically adapted to an inferred level of expertise  using visual cues. In this respect, cogDTs  can recommend actions to steer the course of interactions with virtualized applications toward  successful task completions in the remote real environment. By contrast, conventional robotic hardware with generic interfaces (e.g.,  control panels) can
hardly be modified at this level of reconfiguration and responsiveness. In \textit{Metarobotics}, emerging technologies are leveraged   at consumer levels to provide a pervasive and itinerant  interaction with  remote robotized applications in society, industry, and in-between. Citizens are empowered with the capability to continuously customize the level of control authority $\alpha\in [0,1]$ to shape the  function 
\begin{equation}\label{alpha}
	h(u_h,u_{a})=\alpha u_h+(1-\alpha)u_{a}
\end{equation}
where $h$ arbitrates the autonomy level of the physical distant robot using the citizen input $u_h$ and   autonomous control $u_{a}$, as introduced in \cite{selvaggio2021autonomy}. In \textit{Metarobotics}, however, $u_a=u_{DT}$, which is a cogDT-driven autonomous control. It is worth noting that this choice  injects a holistic  awareness  of society and industry  inferred by interoperating cogDTs in the behavior of the remote local robot. Such an arbitration  is likely to lead to  a  Quality of Value (QoV) that transforms society and industry in terms of \mbox{self-determination,  self-efficacy, and comfort.}

\subsection{Fostering Personal Self-Determination in Society}
  
\begin{figure}[t]
	\includegraphics[width=0.45\columnwidth]{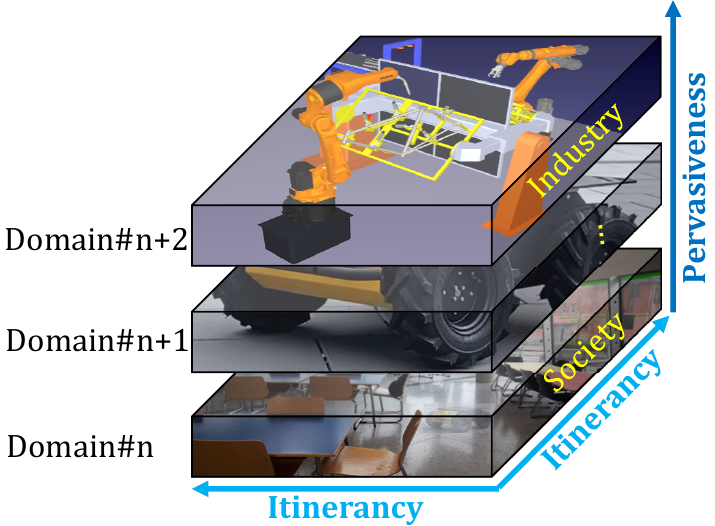}
	\includegraphics[width=0.535\columnwidth]{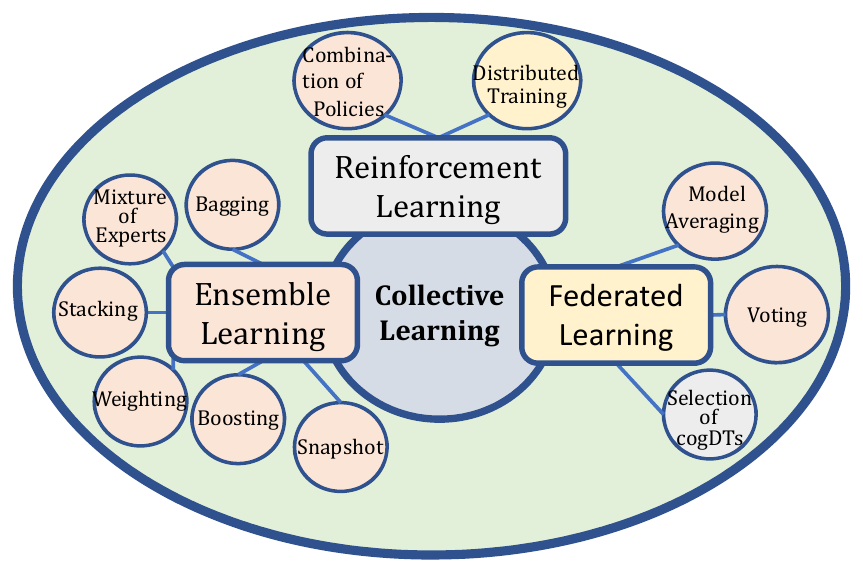}
	\centering
	\caption{\textbf{L.h.s:} Intinerancy and pervasiveness in \textit{Metarobotics}. \textbf{R.h.s:} Three pillars of collective learning in \textit{Metarobotics}.}
	\label{ensemble}
\end{figure}

An itinerant access to virtualized applications allows citizens to spatially self-project from anywhere (see l.h.s. of \cref{ensemble}) to achieve personal objectives   without  being physically present where the robot  carries out domestic tasks in reality. 
Senior citizens, for instance, can thereby be
 physically~\cite{keroglou2023survey} and socially~\cite{abdollahi2022artificial} assisted at home, in supermarkets, or airports. This can be done by tuning $\alpha$ in \cref{alpha} to compensate for a potential disability and preserve autonomy. For $\alpha \rightarrow 1$, a remote family member, as an example, and the local senior   fully influence the robot behavior. A key advantage is that the member can be located anywhere. The immersion is volumetric, i.e., gaze directions, gesture dimensions, and the emotions of the senior  together with  undesired collisions  like accidentally knocking over a bottle (see r.h.s of \cref{metarobotics}) are better perceived than in 2D. A suitable arbitration of the autonomy of the prolonged member' arm (i.e., physical  robot) can also foster self-determination of the senior. Assisting cogDTs in the  Metaverse layer in \cref{threelayer} act on top of the perception layer to automatically adapt the robot behavior  to e.g.  gaze-related \cite{yang2023natural} or speech-based \cite{abdollahi2022artificial}  inferences of needs and intentions of the senior who thereby cogDT-driven ($\alpha\rightarrow 0$) indirectly commands the robot. Since robotic emotional intelligence for social assistance incorporates monitoring, expressing, and understanding emotions~\cite{abdollahi2022artificial}, which are in turn  considerably impacted by cultural factors~\cite{marcos2022emotional,korn2021understanding}, considering  diversity for rich discriminative inferences using  deployed AI$\slash$ML models is important~\cite{fazelpour2022diversity,gong2019diversity}. To this end, Foundation Models broadly and globally pre-trained in the collaboration Metaverse can be specialized to downstream emotion detection tasks through transfer learning  \cite{huang2023study} (see \cref{metarobotics}). Local AI$\slash$ML models (LMs) are trained on-premise using sensitive data of citizens and open  IoT data  to  update  public foundation models (FMs) without disclosing  raw  data via Federated Learning (FL). Scenarios Engineering is a methodology to develop FMs in the Metaverse~\cite{li2022novel}. Resulting Federated Foundation Models (FFM)~\cite{yu2023federated} with relay  mechanisms to support energy-limited  devices~\cite{li2023energy} take advantage of  edge AI for an intelligent (e.g., context- and resource-aware) and continuous access to applications  from society to \mbox{industry with human-centered values.}%

\subsection{Supporting Professional Self-Efficacy in Industry}\label{SEIndustry}
  Orthogonally, pervasiveness describes  the capability of \textit{Metarobotics} to help  robotized applications  penetrate and aggregate different domains of society and industry (see l.h.s. of \cref{ensemble}) as well as forge a collective and shared  generative intelligence that accelerates value creation and uplifts workers (see l.h.s. of \cref{metarobotics}).  For example, customers {and their environments (e.g., a parcel)} can be spatially virtualized using the drone technology and projected into industrial settings for  robotized individualized construction, and conversely. In this regard, the collaboration Metaverse  and mobile Holoportation in \cref{threelayer} are combined  to connect geographically distributed stakeholders (e.g., customers, architects,  mechanical engineers, project managers, etc.) from different disciplines around the joint employment of robots to realize an individual house. Embodied avatars and cogDTs are  drivers of this undertaking.  
  
In \text{volumetrically accessible} \cite{sharmaintuitive} collaboration spaces of \textit{Metarobotics}, cogDTs are aggregated to support two of its core functionalities (see  \cref{metarobotics}) namely itinerancy and pervasiveness (see l.h.s. of \cref{ensemble}). CogDTs transparently  infer contexts, states,  and intentions of stakeholders by e.g. using  parallelizable attention mechanisms of  vision transformers \cite{abdelraouf2022using}   or learning  context relations among  domains for  robust  recognition tasks (e.g., classification, segmentation, detection) despite domain gaps \cite{hoyer2023mic}. Inference results are then used  to adapt the type and scope of created spatial common sense knowledge to enhance the efficacy of workers to safely and comfortably execute tasks \cite{conti2022human} during \textit{itinerancy}. Intersectoral   knowledge is employed on the other side to support \textit{pervasiveness}. Outcomes of  common sense reasoning of interconnected cogDTs finally help stakeholders make efficient decisions. 
 
  In collaboration spaces for robotized  construction, for example,  geometric forms specified by designers are translated into Cartesian motions of the robot end-effector by engineers. A global network of interoperating cogDTs  acts in the background as a parallel intelligence engine (PIE) to quickly optimize  time scaling factors with which joint trajectories are executed with a minimized energy consumption and anticipate bottlenecks. Outcomes of the PIE augment stakeholders from anywhere to validate or reject   how remote robotized processes  semi-autonomously ($\alpha \rightarrow 0.5$) evolve and make informed supervisory decisions for task completion with substantially less resource usage, which increases  the professional satisfaction.

\subsection{Work-Life-Flexibility between Industry and Society}
Giving workers more choice over locations in which they work as a strategy to engage them is a key objective of work-life-flexibility \cite{kossek2023work}. 
In Metarobotics,
the smart mobility of citizens  goes together with their remote multi-modal presence when it comes to e.g. manually guiding a distant robot  in real-time via its cogDT and skillfully supervising or automating in-motion. Resulting advantages are professional and personal. 
Indeed, \text{\textit{Metarobotics}} invites to redefine and individualize the notion of workplace in robotized applications. It shifts workspaces from  restrictive and location-discrete "\textit{office or$\slash$and home}"-rules to  flexible and location-continuous "\textit{from anywhere if desired}"-opportunities in Society 5.0.  
Commuters, such as robot technicians, exposed to project-related traveling, and outsourced employees  can benefit from this shift. They can work from \textit{anywhere}, which impacts their work-life-balance. \textit{Metarobotics} leverages  digital technologies to meet key values like inclusion and co-innovation worldwide. It is thus  likely to not only engage and bind but also  offer opportunities to   the post generation Z to capitalize \mbox{on  curiosity and creativity.} 

                 \begin{figure*}[t!h]
	\includegraphics[width=1.7\columnwidth]{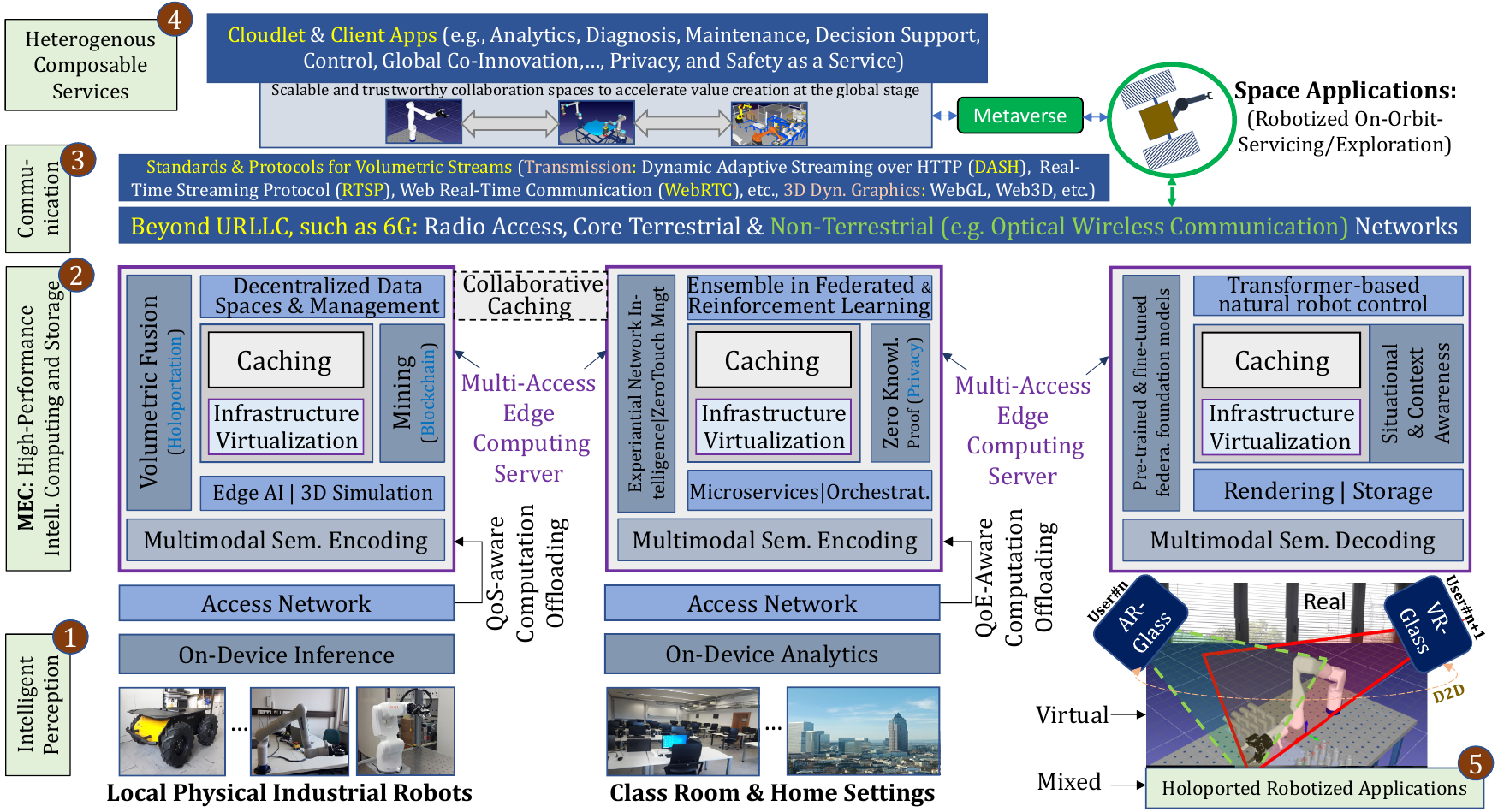}
	\centering
	\caption{
		A high-level architecture of \textit{Metarobotics} with five key layers being highlighted (see numbering). 
}
	\label{overall}
\end{figure*}

\subsection{Targeted Design and Engineering Functionalities}
 \subsubsection{Collective and Generative Knowledge for  Citizens and Robots}\label{knoledge}
 \textit{Metarobotics} makes use of sensing with on-board pre-processing and reasoning based upon  off-loaded AI$\slash$ML at the edge (see \cref{threelayer,overall}). On-device storage and computation limitations as well as  latency     and congestions issues  are  mitigated. 
  Sensitive  data are securely collected  and analyzed through a trusted decentralized AI$\slash$ML-pipeline. It employs loosely-coupled microservices to create new knowledge  (see \cref{overall}). Service discovery and composition, load balancing, circuit breaker function for reliability and fault tolerance, as well as rate limiting against excessive usages need to be implemented. Resilience, adaptation at run-time, and scalability are resulting advantages. FL is run on top of  variants of Ensemble Learning  (see \cref{ensemble}) to address accuracy issues due to significantly heterogeneous class biases among local data~\cite{zeng2022hfedms} while creating FFMs~\cite{yu2023federated}. 
   \textit{Metarobotics}  benefits from FFMs  when AI$\slash$ML models are intended to achieve similar tasks (e.g., picking and placing a rotor in an assembly line and grasping bottles in home settings) using new and legacy  robots with limited data access. Once a FFM is built and shared in the Metaverse, parameter efficient transfer learning (e.g., Adapter, Prompt, Diff-Pruning, BifFit) accelerates the specialization  of the  collective intelligence condensed in pre-trained parameters to downstream tasks. While most parameters are frozen, only a few are fine-tuned, which prevents  building the entire large model from \mbox{scratch and  substantially saves time and energy.} 
   
   A challenge in \textit{Metarobotics} resides in the transferability of trained models from the Metarverse to reality in terms of performance. Even though the Metaverse is fed with live measured data, the cogDT might be model-based and data can be noisy, leading to model uncertainties (e.g., truncation, undesired frequencies). As e.g.  reinforcement learning (RL) is involved in applications, \textit{Metarobotics} can leverage knowledge distillation and reuse \cite{zhu2023transfer}, among others, to address this challenge. In distillation, the  policy in reality is a model that minimizes  action divergences  between Metaverse and reality \cite{zhu2023transfer}. In the reuse case, the real policy is a weighted combination of  policies developed in Metaverse while considering the performance expected from the targeted robot in reality \cite{zhu2023transfer}.
   
   Another challenge in the Metaverse is the design of  modular robots  that can complete a  task  in reality. Being capable to assess the feasibility  of this objective to make decision contributes to self-efficacy introduced in  \cref{SEIndustry}. Given task objectives, a robot configuration can be searched to complete the task with the highest performance. Generative Adversarial Networks (GANs) are used to learn to map one task to a distribution of configurations \cite{hu2022modular}. Motivated by the graph-like kinematics of robots,  a global control policy of  modular robots can be learned using Graph Neural
   Networks (GNNs) in which  knowledge \mbox{is shared among different configurations \cite{whitman2023learning}.}

 Complementary knowledge about entities, such as  physical, contextual,  functional, cultural, and ethical insights,    is targeted in \textit{Metarobotics}. The reason is at least twofold. First, a decentralized semantic storage based upon ontologies for fast and robust queries. Second, the inference of semantic properties of entities under heterogeneous sensory perceptions and task objectives. This is because  symbiotic and thus mutualistic interactions between robots and citizens are envisioned. In this case, inferred knowledge needs to overcome the often handcrafted perspective of citizens~\cite{thosar2021multi,liu2021learning}. Knowledge is therefore also interpreted from the conceptual perspective of individual robots depending upon, for instance, their operational contexts. Unsupervised clustering~\cite{thosar2021multi} or  transformer-based neural networks~\cite{liu2021learning} can be utilized to this end. In combination with human perception and cognition, these  skills strengthen reasoning capabilities of decentralized cogDTs. Autonomous inference and substitution of missed information in remote applications \textit{"from household to industrial robotics"}~\cite{thosar2021multi} even in deep space applications by combining terrestrial and non-terresrial networks are {targeted outcomes of \textit{Metarobotics}, as depicted in \cref{overall}).

 \subsubsection{CogDT-Based Multi-Agent Optimization}
 Swarms of distributed cogDTs  are assumed to operate as systems of decentralized optimization agents in \textit{Metarobotics}.  Each agent individually explores a solution space  and  shares its experience to collectively contribute to the optimization of a cost function to complete robotized tasks. Semantics-driven M2M-communication standards, such as OPC UA FX, can support  this collaboration. Related information models enable a common  and global understanding of  exchanged data, helping out each cogDT to self-adapt to cumulative broadcasts of  findings sent by other cogDTs. Similarly to the particle Swarm Optimization (PSO), the approach is gradient-free. It is not undermined by discontinuous cost functions.  However, in contrast to the  PSO, advantages of   CogDT Swarm Optimization (CSO)  include the  heterogeneous abstraction (e.g., modular robots with distinct configurations designed as in \cref{knoledge}) and reasoning capability of each cogDT. Instead of uniformly enforcing  predefined behavior patterns  for each  swarm member as in the PSO, each cogDT independently  exploits  attention-based rewards to contextually and situatively reasons and decides \mbox{how to sustainably contribute to swarm objectives.}

 \subsubsection{Green Ultra-Massively Adopted  $\pi$-HRC}
Surroundings- and application-awareness for b-URLLC-based mobile Holoportation (see \cref{holoportation}) can be used in \text{\textit{Metarobotics}} to democratize the sustainable access to remotely conducted robotized applications. Mobile Holoportation has been initiated by Microsoft in the automotive field and remains an active research field \cite{holoportationMicrosft}. In the robotics context of \text{\textit{Metarobotics}}, the adoption of mobile Holoportation by an ultra-massive number of geographically distributed stakeholders  is driven by two incentives in the Society 5.0 context.  First, an individualized QoE of citizens that also aligns with objectives of Industry 4.0.  Needs and intentions of workers and robots are inferred to enhance process efficiency through anticipation. High resolution sensing allowed by the Teraherz and  mmWave spectrum \cite{chen2022tutorial,wymeersch2020radio} and collective localization of physical siblings are enablers. To this end, AI$\slash$ML, such as RL, can  scrutinize emitted and acknowledged sensing data to capture a channel model for motion inference and thus localization \cite{chen2022tutorial,wymeersch2020radio} of physical siblings of collaborating decentralized cogDTs. 6G-based centimeter-level  3D localization (see \cref{table:v4v5}) helps understand root causes (e.g., obstacles) and prevent collisions of mobile robots in holoported spaces in real-time. Second, the  QoV like  work-life-flexibility stemming from decent work conditions and multi-stage characteristics of the underlying sustainability as advocated by Industry 5.0.  %
In \textit{Metarobotics}, the inherent energy efficiency of smart mobility is combined with minimized  energy consumption of 6G  (see \cref{table:v4v5}) together with an active reduction  of  energy expenditures of  robots. Remote time scaling of joint trajectories or posture optimization in the Jacobian null-space while productively meeting  goals with the decoupled \mbox{end-effector can be done to this end.}

 \subsubsection{Holoportation with  Sensation Feedback}\label{multi-modalholoportation}
 Holoportation with sense of touch   is expected to be pivotal in \textit{Metarobotics}.  It complements visual immersion and  enhances the QoE and QoV. For instance, being aware and feeling the exchange of mechanical energy with remote entities 
 meet various goals. In the former case, force feedback indicates  physical interactions with other entities. In the latter case,  tactile sensations  come  into play. A high resolution of measuring  how pressure is distributed over contact areas is enabled.  Tactile sensations thus provide a more accurate perception of distant and  occluded objects in even  cluttered areas. Touch-sensitive applications, such as tactile manual guidance in $\pi$-HRC and remote lights-out manufacturing, benefit from this perception of entity properties. These include   stiffness and softness (e.g., elasticity, plasticity), roughness, geometry, and contact force localization. %
 
Motivated by ISO/IEC TS 23884:2021(E), at least three approaches can  help realize the sense of touch in AR$\slash$VR-based interactions in \textit{Metarobotics}. The first one, used in advanced multi-body simulation, exploits parameters for constraint force mixing and error reduction  to mimic customizable spring and damping behavior of entity materials during contact dynamics.  The second approach is a real-time simulation of an elastic tactile sensor as proposed in~\cite{wang2021elastic}. In this case, the tactile sensor is voxelized to yield particles as voxel centers. Their displacements  under pressure help render deformation processes by using the (Moving Least Squares) Material Point Method~\cite{wang2021elastic}. The deformed mesh is then reconstructed on the basis of the particle location~\cite{wang2021elastic}. An advantage of this second method for \textit{Metarobotics} is the  dynamic-visual and tactile information perceived by citizens. This enriches their  immersion  in \mbox{\textit{Metarobotics} especially when soft materials are considered.} 
 
 Since human skin and grasped objects, including a robot, might have varying temperatures, as in glass factories~\cite{barker2022collaborative}, thermo-tactile feedback is considered in the third approach \cite{sun2022augmented}. A  triboelectric and pyroelectric ring-sensor  worn by a citizen and connected to the holoported environment via 6G  reflects the thermo-tactil feedback. A  nanogenerator  tracks how muscles  swell when fingers bend to estimate  pressure by integrating voltage~\cite{sun2022augmented}. A heated nichrome
 metal wire provides thermo feedback in the ring~\cite{sun2022augmented}. Thermo-tactile feedback fosters values of Industry 5.0 and Society 5.0 in \textit{Metarobotics}, such as socialization,  resilience, and diversity. Indeed, geographically distributed collaborators can showcase and familiarize with specific cultural rules of People-to-People (P2P) communication and etiquette, including remote hugs and handshakes in meetings~\cite{sun2022augmented}, \mbox{without e.g. contamination hazards.}

  \subsection{Parallel Intelligence for Human-Centered  Robotics}\label{parralelIntelligence}
 Embedding visual and tactile features in a latent space can help cogDTs predict tactile forces from images \cite{takahashi2019deep}. 
Learned multi-modal correlations between images and tactile features allow it to  adapt robots  and inform citizens about various environment properties like roughness \cite{takahashi2019deep}. Cross-modal reasoning can be achieved using interconnected   multi-modal knowledge graphs (MMKG)~\cite{zhu2022multi,li2022novel} that evolve from structured crowd sourcing, IoT, and synthetic data. 
 This leads to a swarm of cogDTs (as MMKG-nodes) with maturing intelligence in terms of the depth and diversity of knowledge integration, as well as  edge and cloudlet  processing rates (see \cref{overall}). The swarm  acts as a PIE alongside and beyond human capabilities. Experience is captured  by e.g. grounding representation symbols to their semantics in the real word structured in the MMKG~\cite{zhu2022multi} to empower industry and society. In \textit{Metarobotics}, the symbol robot can be grounded to  contextual and situational multi-modal data like joint state, ML-models, videos, and CAD files to \mbox{enlarge the  breadth of experience.}
 
 \subsection{Global, Trustworthy, and Cross-Domain Robotics}
   \textit{Metarobotics} aims to capitalize on several opportunities delivered by the Metaverse to  lower entry barriers and revamp  the collaborative realization of distant robotized applications. Its location-agnostic accessibility can yield an engaging  proximity, visibility, democratization, experimentation, and familiarization with robotics. In the Metaverse, which is also viewed as an interconnection of  decentralized virtual collaboration spaces loosely coupled with digital and physical assets, as shown in \cref{robotaas}, robotics-related contents in terms of knowledge, service, and products are globally created and  exposed as well as instantly discovered, purchased, and combined. Contents are consumed by cogDTs of assets, processes,  and citizens worldwide using  Blockchain-based Non-Fungible Tokens (NFTs) for e.g. authenticity, authorization, and ownership check. \textit{Metarobotics} aims to enable a sovereign data sharing to preserve privacy. Following the  connector idea of the International Data Space Association and project GAIA-X \cite{otto2021europaische}, \textit{Cotrusting} can manage who is granted access to raw data by initially exposing only meta-data instead of raw data. Upon agreement, encrypted raw data are sent to authorized cogDTs. Raw data are  jointly processed with trust by design (see \cref{selforga}).  Virtualized robotized resources and services are combined through standardized interfaces for enhanced performances. They are then deployed in  collaboration spaces to quickly and cost-effectively assess and shape benefits of robotics in untapped markets. Resources are parts of a global society and circular economy (based upon interoperable NFTs) in the Metaverse. While constraints and shocks can be virtually customized, interactions with physical distant assets use measured data, raising the \mbox{applicability of assessment results.} %

   \subsection{Streamlining Education Worldwide}
     Learners can expect elevated experience in robotized applications under authentic   industrial or societal   conditions close up with \text{\textit{Metarobotics}} (see \cref{robotaas}). These include  learning best practices even in the early innovation phases. \textit{Humotics} allows teachers to extend  excursions often  restricted to local factories, for e.g. logistic reasons, to concerns abroad at negligible complexity and costs. Since an intrinsically safe, volumetric, and multi-modal proximity to distant physical robots will be ensured in \textit{Metarobotics} by  exposing learners to  cogDTs, tremendous intrinsically safe possibilities for course design arise.  
     Immersive learning is customized to the individual skills and background of learners and teachers using certificates. Learners  discuss ideas with on-site  experts in real-time to validate theoretical results. Industrial and societal open  data are provided to laboratories. Students thereby learn with authentic data how to transform industrial resources into competitive advantages using AI$\slash$ML-based analytics. Developed prototypes, such as filters and pre-trained models, are re-injected in the Metaverse to \mbox{cross-fertilize  industry, society, and academia.}

      \section{Architecture and Enabling Technologies}\label{enablingtechnologies}
      An architecture of \textit{Metarobotics} is given in \cref{overall}. Enabler technologies  therein  provide together functionalities  to fulfill goals stipulated in \cref{Metarobotics}. Data are collected in the first perception layer. Encrypted relevant results of on-device pre-processing are  offloaded to the second microservice-enabled MEC-layer for further processing. The third communication layer is b-URLLC-driven. Delay- \cite{rodrigues2019edge} and Energy-efficient \cite{dou2023energy} cloudlets can be  deployed in the fourth layer to benefit from  edge proximity and  resources elasticity. This streamlines the execution of  $\pi$-HRC applications \mbox{in the final fifth layer.}
      \subsection{Communication Driven by  b-URLLC}
    \subsubsection{Tactile Internet}
    Interactions  between citizens  and remote  robotized applications involve tactile sensors with spatial resolutions resp. sampling rates below $0.5$~mm resp. above $10$~KHz. 6G-based tactile internet is expected to meet global network connectivity needs for a massive number of industrial and consumer-level applications.   Ultra-reliable data transmission rates in multiple TBps with a stringent sub-millisecond latency (see  \cref{table:v4v5}), i.e., below the time of reaction of citizens ($\approx$0.2s), and substantially reduced latency jitters and packet losses will improve the QoS and QoE in \textit{Metarobotics}. Indeed, delays raise the sensation of heavier objects and jitters induce not only instabilities, but also  the misleading sensation of a varying mass of objects~\cite{marshall2008providing}. An accurate perception of the Cartesian effective mass of robots in given directions is however safety-relevant in robotized applications, such as physical $\pi$-HRC. Since packet losses {distort the power of the perceived force~\cite{marshall2008providing}, \textit{Metarobotics} will benefit from the enhanced reliability of 6G  wireless communication and beyond, as quantified in  \cref{table:v4v5}.} Nevertheless, data synchronization in volumetric fusion and coordinating different signals such as  visual, audio, thermal, and haptic data streams with distinct latency values require  \mbox{more attention to further enhance the QoE and QoS.}
     
    \subsubsection{Haptic Feedback Support}
    In \cite{xu2022timestamp}, the transmission of a haptic-visual signal that does not depend upon timestamp synchronization has been proposed. The synchronization is instead based upon the combination of key samples of haptic and key frames of visual signals by taking advantage of the sequential correlation observed in the transmission and playback~\cite{xu2022timestamp}. Context-aware haptic feedback is addressed in~\cite{mondal2020enabling} by adopting a two stage approach. First, a supervised learning that relies upon Artificial Neural Network (ANN) is applied to control data from a VR glove and predict whether haptic feedback is necessary or not with an accuracy of $99\%$. Then, RL is used to predict samples of haptic feedback with an accuracy of $92\%$ for four different materials~\cite{mondal2020enabling}. Characteristics of People to Machine (P2M) traffic in haptic teleoperation, such as the packet interarrival time and the correlation between human control and haptic feedback during a time window equivalent to a polling cycle, are estimated in \cite{ruan2021achieving}. Whereas the generalized Pareto arrival model provides  the smallest fitting error  when compared with three other statistical distributions (t-location, Logistic, and
    Exponential),   a considerable cross-correlation between 0.6 and 0.8 is observed between control (master to slave) and feedback (slave to master) traces.  Therefore, an ANN estimates  bandwidth requirements for P2M traffics. The bandwidth is then  predictively allocated to the control traffic and, at the same time,  interactively  granted (by harnessing the correlation) to the feedback traffic to reduce latency and  accelerate haptic feedback  \cite{ruan2021achieving}. In the multi-modal case (e.g., visual, haptic, and audio signals), haptics-related control traffic is allocated additional bandwidth unlike content requests. Furthermore, haptic feedback packages are assigned a higher priority \mbox{during the transmission for a lower latency \cite{ruan2021achieving}.}

    \subsubsection{High Mobility Support}
    Intelligent 6G networks are expected to learn and predict properties and communication requirements of applications to adapt their operational configurations and  enable multi-modal   tactile communication capabilities under high mobility~\cite{fettweis20216g}. %
    Predictions can result from historical observations when it comes to e.g. train beams for the use of mmWave and THz bands to support  increased mobility \cite{hou2021intelligent}. Online AI$\slash$ML, such as RL, operates at the network edge to optimize the usage of resources  via efficient anticipations of upcoming demands that leverage e.g. Open Source Multi-access Edge Computing (OS-MEC)~\cite{zhao2021open}  to fulfill  constraints on transmission delay~\cite{strinati20216g}. OS-MEC ensures dis-aggregation, i.e.,  a separation and adaptation of   MEC functions and resources, for a flexibly tailored edge performance by taking advantage of Network Function Virtualization (NFV)~\cite{zhao2021open}. End-to-end  delays include not only the elapsed time for caching, computing, and transmission~\cite{strinati20216g}, but also the overhead to train the beam to estimate  channel states \cite{hou2021intelligent}.  A GNN that generalizes over network structure, routing approach, and traffic intensity and predicts the average delay and jitter is combined with RL  to  optimize routing strategy  and congestion control via Software Defined Networking (SDN) in~\cite{hou2021intelligent,rusek2020routenet}. Furthermore, a mobility management that foresees the  cell in which a citizen  in motion is likely to enter and be allocated radio  before handover is proposed by \cite{hou2021intelligent}. This prediction  helps  reduce the overhead of signaling\cite{hou2021intelligent} along with related latency and power consumed to this end. It also achieves a continuous coverage during mobility \cite{hou2021intelligent} and {thus enhance QoS and QoE in \textit{Metarobotics}.} In addition to a low power consumption of 1 pJ/bit for a sustainable wireless communication, as mentioned in  \cref{table:v4v5},  mobility support at a speed of up to 1000 Km/h is a prominent advantage for sustainable pervasiveness \mbox{and itinerancy goals pursued by \textit{Metarobotics}.}

	\subsection{Cognitive Digital Twin}\label{cohgnitionDTW}
	
	\subsubsection{Cognition}\label{fromdttocogdt1}

Industrial and personal applications are  subject to 
uncertainties. Noisy data, truncated models, faults, anomalies, and undesired contact forces are a few  causes. For the sake of adaptable and robust interactions between citizens, cogDTs, and physical assets, the standard DT is equipped in \textit{Metarobotics} with  cognitive  skills. These encompass \textit{"perception"},  \textit{"attention"}, and \textit{"reasoning"}~\cite{mortlock2021graph}. Perception aims to a meaningful representation of  sensed and accessible data about entities~\cite{mortlock2021graph}. Attention supports the selective concentration on relevant information~\cite{mortlock2021graph}. To create knowledge and anticipate  uncertainties,  knowledge engines of cogDTs  learn on data structured in a knowledge graph (KG). These engines usually combine sub-symbolic
(e.g., Multi-Layer Perceptron (MLP), Neural Tensor Network (NTN), Deep Learning) with symbolic (e.g., ontologies-, rules-, and
expert systems-based) AI$\slash$ML,  to predict properties   and missing relations (i.e., edges) between entities (i.e., nodes) of a KG as well as their clustering and constraints. In \textit{Metarobotics}, MLP and NTN, for instance, can be used to  capture correlations between nodes and edges by learning latent features when properties cannot be directly observed~\cite{DBLP:journals/corr/SaxenaJSJMK14,nickel2015review}. Since ontologies like OWL meaningfully formalize relations, rules, and constraints between nodes, they are  interpretable  by cogDTs, robots, and citizens. This in turn supports  M2M, P2M, P2P, as well as cogDT to cogDT, Machine to cogDT, and Citizens to cogDT communication in \textit{Metarobotics}. Further types of functional constraints along with incompatibilities can be learned by observing sets of nodes and edges~\cite{nickel2015review}.   Hence,
a skillful, fast-growing, and evolving ecosystem with an actively harnessed  latent \textit{"body of knowledge"}~\cite{kinsner2021digital} and  \textit{"body of experience}~\cite{kinsner2021digital}
arises, in which cogDTs find inputs about what they
reason. The goal of the  cogDT-based PIE is to predict and anticipate events, while defining and
scheduling the next actions to be recommended to citizens and robots in \textit{Metarobotics}, as highlighted in \cref{robotaas}. 

\subsubsection{Uncertain Knowledge Graph}\label{uncertainKG}
Uncertainty accommodation will be essential to handle  unfamiliar and  unforeseen operational conditions in remote workspace in \textit{Metarobotics}. CogDTs  generalize knowledge from known facts to infer  course of action and adapt.
For instance, a cogDT can  contextualize and characterize an initially unknown object  as a  grasp target of the robot (see r.h.s of 	\cref{metarobotics}). 
The trustworthiness of relations between a payload and  similar workpieces previously manipulated are used to this end. Therefore, relations between pairs of KG nodes are assigned  confidence scores that  reflect the level of belief in the relation to take uncertainties  into consideration~\cite{chen2019embedding}. This contrasts with deterministic KG in Robobrain~\cite{DBLP:journals/corr/SaxenaJSJMK14}, where the belief is maximal. The capability  cogDTs to transfer knowledge can be based upon the classification of  nodes and prediction of links between them~\cite{nickel2015review}. A knowledge engine can thereby  discover and recommend facts in addition to efficiently  answering  queries for even unseen facts~\cite{chen2019embedding}. Nodes and relations of such an uncertain KG are embedded into a low-dimensional continuous (latent) space~\cite{chen2019embedding}. Efficiency refers to insightful abstractions of  non-Euclidean and multi-modal data  as well as the richness and expressiveness of   representations learned using e.g. GNN~\cite{mortlock2021graph}. Probabilistic soft logic and probabilistic box can be used to  predict  confidence scores of unseen facts by transferring  confidence scores \mbox{related to available knowledge to unseen relations~\cite{chen2019embedding}.}

  \begin{figure}[t]
	\includegraphics[width=\columnwidth]{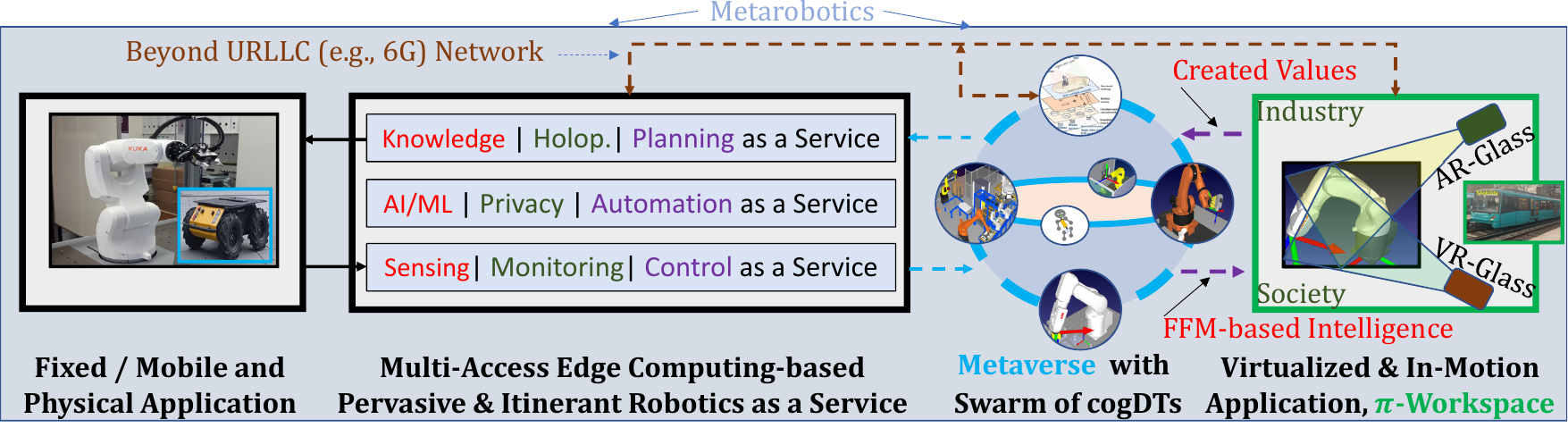}
	\centering
	\caption{Metaverse is service$\slash$content provider for \textit{Metarobotics}.}
	\label{robotaas}
\end{figure}

\subsection{Holoportation}\label{holoportation}
The pipeline for Holoportation includes   scene capture,  volumetric fusion, transmission, and rendering phase \cite{orts2016holoportation,langa2022multiparty}. 

\subsubsection{Capture}
Surrounding RGB-D cameras like Kinect can be used to capture scene from which point clouds (PCs)~\cite{langa2022multiparty,viola2023volumetric} or time-varying 3D mesh (TVM) \cite{orts2016holoportation,viola2023volumetric} are generated.

\subsubsection{Fusion} For PCs, a synchronized merging of RGB-D frames can be conducted for reconstruction~\cite{montagud2022towards}.
In the TVM case, a deformation model  of the nonrigid motion field between  frames can be used for a temporal volumetric fusion from which a 3D polygonal model is obtained \cite{orts2016holoportation}. 
\subsubsection{Encoding and Transmission}
The compression aims to balance real-time capability, low-latency, and quality, which aligns with a high QoS and QoE.
In the PCs case, octree occupancy can be used to represent geometry~\cite{langa2022multiparty}. Vertex de-duplication, the reduction of position and normal data, and the assignment of a constant color to non-foreground from  segmentation help reduce the frame size in the TVM case \cite{orts2016holoportation}. Dynamic Adaptive Streaming
over HTTP (DASH) currently support both mesh- and point cloud-based Holoportation \cite{viola2023volumetric}.

\subsubsection{Latency and Throughput Challenges} 
Holoportation throughputs are in the  Gbps range and  already supported by 5G networks with a 4K spatial light modulator for displaying~\cite{he2023three}. However, in VR, the experienced computation and communication latency of more than $140$~ms is at least $9\times$  the maximally allowed latency (< 15 ms)~\cite{elbamby2018toward}. Also, ultra-high definition in 8K with e.g. 48 Gbps is better supported by 6G  (see \cref{table:v4v5}). Computation resp. communication latencies can be reduced using MEC~\cite{elbamby2018toward}. In \textit{Metarobotics}, robot abstractions can help further reduce latency if e.g. the environment of the remote robot does not change. Without high dynamic modes,  the robot posture is retrieved from its forward kinematics and  rendered on the receiver side. Only current joint positions (i.e., a vector of  scalars) are therefore transmitted without expensive point cloud processing. As dynamics are more involved,  kinematic and dynamic models can be learned by using GANs \cite{ren2020learning} 
and retrieved once joint positions {and velocities of the remote robot are received.} Conversely, the velocity of external joint torques induced by  manual guidance can reveal citizen \mbox{intentions to manually accelerate or decelerate  robots.} 

\subsection{Metaverse}\label{Metaversesec}
Another key enabler of \textit{Metarobotics} is the Metaverse, as shown in \cref{robotaas}. It is a digital ecosystem fed with hybrid (e.g., sensed,  synthesized) data and  populated by cogDTs that mirror real and prospective applications. Usually, this occurs in decentralized and interconnected  virtual collaboration spaces.  \textit{Metarobotics} projects these spaces onto further digital and physical environments (e.g. shop floor, home settings, trains) on-demand  and re-injects experience, knowledge, and wisdom  gained from completing robotized applications back into the Metaverse for efficient cross-fertilization and prosperity. This bidirectional communication is depicted in \cref{robotaas}. 

The \textit{Omniverse} platform~\cite{nvidiabmwmeta}, recently released by Nvidia, is increasingly in use at e.g. BMW \cite{nvidiabmwmetayoutube} and Ericsson \cite{nvidiaericson} to develop such places in the manufacturing and telecommunication realm. In \textit{Metarobotics}, however, citizens are empowered with tools to influence the course of action of robotized applications in society and industry as well, adhering to the concept of metasocieties that extends real society with skillful and far-reaching forecasts and suggestions \cite{wang2022metasocieties}. \textit{Metarobotics} targets a global, standardized, and trusted robotics-related approach that leverages AI$\slash$ML-based emerging technologies, such as FFMs, to  accelerate \mbox{familiarization, adaptation, and self-fulfillment.}  

\subsubsection{Interoperability} A  mutual  understanding of formats  for scene modeling, processes, and services is pivotal to use cogDTs and enhance QoE across collaboration spaces involving heterogeneous tools. ISO/IEC 23005 standardizes interoperability between collaboration spaces as well as physical and collaboration spaces~\cite{ISOIEC23005}. It provides an architecture and information models for data traffics and specifies data formats for e.g. robotic devices, such as sensors, actuators, and virtual assets~\cite{ISOIEC23005}. ISO/IEC 23005 considers the use of virtual decisions to command \mbox{physical assets, as pursued by \textit{Metarobotics}.}

\subsubsection{Self-Organization, Privacy, and trust}\label{selforga}

Blockchains  offer a  tamper-proof collective storage and synchronization for self-organization in \textit{Metarobotics}. Traceable  and encrypted transactions  can be automatically triggered and executed as blockchained smart contracts  in  decentralized networks of cogDTs using a bi-level coordination for resilience and autonomy \cite{leng2023manuchain}. The authenticity  of data  and services provided by  cogDTs can be verified using Self-Sovereign Identity with Zero Knowledge Proofs (at MEC in \textit{Metarobotics}, see  \cref{overall})  without  data traffics over internet or server storage \cite{ghirmai2022self}. This implies that, for \textit{Metarobotics},  citizens and cogDTs can share the same authentication for distinct collaboration spaces \cite{ghirmai2022self}. As trustworthiness  with chemical signature (e.g., in 3D printing) {is needed, makerchains help against counterfeiting~\cite{xiong2022survey}.}

\section{Conclusion}\label{conclusioon}
\textit{Metarobotics} targets a collectively informed usage of remote robots operating in different environments from anywhere. It leverages emerging technologies for a trustworthy, pervasive, and itinerant access to and interaction with remote robotized applications. This paper has surveyed relevant technologies toward this end, such as cogDTs, 6G, Holoportation, and Blockchain. It has also highlighted their  integration based on microservices, dynamic interplay, and usefulness to meet goals of \textit{Metarobotics}. How the QoE and QoV in professional and personal activities can be elevated, has been introduced. 

Challenges remain to be  addressed. Assessing the performance of mesh- and point cloud-based Holoportation in contact and non-contact robotized applications deserves further investigations. Standardizing  a dedicated and ressource-adaptive protocol for Holoportation is likely to strengthen interoperability and consumer-level adoption. In this regard, \textit{Metarobotics} will benefit from developing transmitter and receiver  chipsets for 6G beyond the 100 GhZ \cite{gustavsson2021implementation} and their massive societal and industrial penetration.  
A realistic embodiment of avatars for acceptance and QoE purposes requires contextually inferred  appearances. An approach to realize this objective could be the image-based representation from ISO/IEC 23488:2022(E). Furthermore,  multi-modal feedback, locomotion, and gestures  need to be synchronized with  subjective   gaze-related emotions. 
Although \textit{Metarobotics} targets robotics, it is transferable to other domains. Pervasive and itinerant Product Lifecycle Management ($\pi$-PLM), regardless of the product, is an example.


\begin{spacing}{0.95}
\bibliographystyle{ieeetran}
\bibliography{./myref}
\end{spacing}

\end{document}